\documentclass[review]{elsarticle}

\usepackage{lineno,hyperref}
\usepackage[numbers]{natbib}
\usepackage{enumerate}
\usepackage{enumitem}
\usepackage{amsmath}
\usepackage{xcolor}
\usepackage{soul}
\usepackage[justification=centering]{caption}
\DeclareMathOperator{\atantwo}{atan2}

\DeclareMathOperator{\asin}{asin}
\modulolinenumbers[5]
\usepackage{multirow}
\usepackage{graphicx}
\usepackage{dirtytalk}

\journal{Journal of \LaTeX\ Templates}

\begin{document}

\begin{frontmatter}

\title{Geolocation estimation of target vehicles using  image processing and geometric computation}

\author[mymainaddress]{Elnaz Namazi\corref{mycorrespondingauthor}}
\cortext[mycorrespondingauthor]{Corresponding author}

\author[mymainaddress]{Rudolf Mester}

\author[mysecondaryaddress]{Chaoru Lu}

\author[mymainaddress]{Jingyue Li}

\address[mymainaddress]{Department of Computer Science, Norwegian University of Science and Technology (NTNU), Trondheim, Norway}

\address[mysecondaryaddress]{Department of Civil Engineering and Energy Technology, Oslo Metropolitan University (OsloMet), Oslo, Norway}

\begin{abstract}
Estimating vehicles' locations is one of the key components in intelligent traffic management systems (ITMSs) for increasing traffic scene awareness. Traditionally, stationary sensors have been employed in this regard. The development of advanced sensing and communication technologies on modern vehicles (MVs) makes it feasible to use such vehicles as mobile sensors to estimate the traffic data of observed vehicles. This study aims to explore the capabilities of a monocular camera mounted on an MV in order to estimate the geolocation of the observed vehicle in a global positioning system (GPS) coordinate system. We proposed a new methodology by integrating deep learning, image processing, and geometric computation to address the observed-vehicle localization problem.  To evaluate our proposed methodology, we developed new algorithms and tested them using real-world traffic data. The results indicated that our proposed methodology and algorithms could effectively estimate the observed vehicle's latitude and longitude dynamically.

\end{abstract}

\begin{keyword}
Target-vehicle localization \sep vision-based localization \sep geometric computation \sep GPS coordinate system \sep mixed traffic
\end{keyword}

\end{frontmatter}


\section{Introduction}

A growing body of literature recognizes the importance of intelligent traffic management systems (ITMSs) to manage traffic safely and efficiently. ITMSs mainly rely on traffic data to enhance traffic scene awareness and make smart decisions \cite{namazi2019slr}. 

There are two main approaches to collecting traffic data for ITMSs. The first approach is based on stationary sensors placed toward road networks, such as inductive loop detectors (e.g., \cite{oliveira2010vehicle}) and closed-circuit television cameras (e.g., \cite{kurdi2014review}). Although this approach is nowadays widely applied to collect traffic data, installing and maintaining these sensors to provide an acceptable coverage range on all roads might be costly \cite{coststatistics}. The second approach is based on using modern vehicles (MVs) equipped with sensors. An MV with sensing and communication abilities can collect traffic data mainly about itself and transfer it based on, in general, vehicle-to-vehicle (V2V) or vehicle-to-infrastructure (V2I) communications. To collect enough traffic data with this approach, most vehicles in the traffic need to be MVs with an advanced sensor mounted. However, converting most vehicles into MVs is time-consuming. Studies predict that only 50\% of vehicles in the United States will have autonomy in Level 4 (vehicles in Level 4, based on the Society of Automotive Engineers (SAE), have high automation, with which the automated driving features can drive the vehicle under limited conditions, and the driver holds control only if the automated situation turns unsafe  \cite{ahangar2021survey}\cite{SAE}) by 2050 \cite{raj2020multicriteria}. Thus, the near-future traffic would be a mixture of human-driven vehicles (HDVs) and MVs with various levels of sensing capabilities, which is called mixed traffic hereafter. Therefore, it is necessary to explore the possibility of using an MV equipped with a low-cost and popular sensor (e.g., a monocular camera), with the purpose of enhancing generalizability in mixed traffic to collect traffic data of the observed vehicles and feed them into the ITMSs. 

In our previous studies \cite{namazi2019master}\cite{namazi2020lane}, we have investigated the feasibility of using a vehicle equipped with a low-cost front-facing monocular camera with a built-in global positioning system (GPS) receiver (hereafter, we call this vehicle an ego vehicle) to observe another vehicle  (hereafter, we call this vehicle an target vehicle) and estimate its speed, distance, and the lane it is in. After studies \cite{namazi2019master}\cite{namazi2020lane}, a follow-up research question has been raised about the use of an ego vehicle in estimating the geolocation of the target vehicle, as accessing the vehicle's geolocation plays a critical role in modeling the traffic scene and making smart decisions by ITMSs. 

Therefore, in this paper, we go beyond the lane-level target-vehicle localization presented in \cite{namazi2020lane} and find the latitude and longitude of a target vehicle in a GPS coordinate system dynamically while both the ego vehicle and the target vehicle are moving in a metropolitan area. Although some research has been carried out on utilizing an ego vehicle as a mobile sensor to estimate traffic data of the target vehicle, there is still very little scientific understanding of estimating the geolocation of HDVs based on ego-vehicle self-localization, image-based estimated distance to the target vehicle, and the relative angle between them by using a monocular camera with a built-in GPS receiver mounted on a mobile ego vehicle.

Therefore, the objective of this paper is to investigate the feasibility of using data from low-cost sensors (i.e., a monocular camera with a built-in GPS receiver) mounted on an ego vehicle to estimate the geolocation of a moving target vehicle. Our research question is defined as follows:

\begin{itemize}
    \item RQ: How can the geolocation of a mobile target vehicle be dynamically estimated in a GPS coordinate system based on the vision of a front-facing low-cost monocular camera with a built-in GPS receiver on a mobile ego vehicle?
\end{itemize}

To address this research question, we proposed two approaches based on (1) object detection and image processing and (2) geometric computation by considering the camera's pitch angle and height from the road surface. In this regard, we extended the proposed algorithms presented in \cite{namazi2020lane} by including the estimation of the distance and angle between the ego vehicle and the target vehicle. 

To evaluate our proposed approaches and develop algorithms, we ran empirical experiments using real traffic data from a metropolitan area in Chengdu, China. We analyzed the findings by plotting the estimated target vehicle's trajectory on Google Maps and compared it with the ground-truth trajectory of the target vehicle. Additionally, the vector distance was used to quantitatively analyze the deviations between the estimated and the ground-truth geolocations of the target vehicle. The evaluation results confirmed that both approaches could estimate the geolocation of the target vehicles accurately.

The rest of the paper is organized as follows. Section 2 gives a brief overview of the recent history related to vehicle localization approaches. Section 3 explains the research strategy and methodology we propose. Section 4 presents the experiments and results of our approaches. The discussion is presented in Section 5. The last section concludes and proposes future studies. 

\section{Related work}

To estimate the target vehicle's geolocation in a GPS coordinate system, we need to know the ego vehicle's geolocation and the target vehicle's location (the distance and angle between the ego vehicle and the target vehicle) \cite{distance}. This section presents a related work of these aspects briefly.

\subsection{Ego-vehicle geolocation}
There has been an increasing amount of literature on estimating the geolocation of ego vehicles, which is usually called self-localization (e.g., \cite{javanmardi2019autonomous}). 
A GPS receiver is one of the most popular sensors for localizing ego vehicles \cite{chehri2019survey}. 
Standard GPS receivers in the market have an accuracy of about 10-15 meters in 95\% of the time \cite{gpsac}. To minimize the GPS receiver's estimation error in ego-vehicle localization, map matching is applied widely \cite{huang2021survey}. Huang et al,  \cite{huang2021survey} classified the map matching algorithms into four categories: geometric theory, topology, probability statics, and advanced model.

\subsection{Target vehicle's location estimation}

To date, various studies have investigated target vehicle's location estimation via monocular cameras regarding driving safety measures, assistance, and autonomous navigation. For instance, Ifthekhar et al.,  \cite{ifthekhar2015stereo} introduced an optical camera communications (OCC)-based cooperative vehicle positioning (CVP) technique. They proposed two approaches: (1) a neural network-based approach and (2) a computer vision-based approach to estimate the target vehicle's location. They considered two vehicles, one as an observing vehicle equipped with front-left and front-right cameras. Another vehicle was treated as a target vehicle, and its positioning was estimated based on its rear light-emitting diodes (LEDs). Simulation results showed that the accuracy achieved by the proposed neural network-based method was higher than the computer vision-based method \cite{ifthekhar2015stereo}. Hayakawa et al., in \cite{hayakawa2019ego} proposed a new approach based on integrating three deep neural networks to estimate the ego-motion and the target vehicle's state (e.g., 3D vehicle bounding box, depth, and optical flow). The experimental
evaluations demonstrated that the distance error in the lateral and longitudinal directions were 1.19 m and 1.70 m, respectively.

Lee  \cite{lee2018intervehicle} focused on inter-vehicle distance estimate based on lane width. The proposed technique had a distance estimate error of less than 7\%. Huang et al. \cite{huang2017vehicle} proposed a novel approach to estimate the inter-vehicle distance based on vanishing point detection, road segmentation, and vehicle detection. The ratio of true distance to image pixel was used to calculate the distance. In \cite{huang2017vehicle}, a single-lens camera was utilized to capture data from urban/suburban roadways. Five image sequences of urban/suburban roads were utilized to verify the performance of the suggested method. The results showed average detection rate (DR), and false alarm rate (FAR) values of the approach are 82.21\% and 16.16\%, respectively \cite{huang2017vehicle}. Giesbrecht et al.  \cite{giesbrecht2009vision}  proposed a vision-based leader/follower system for an ego vehicle. The system was a combination of three main components: (1) a computer vision system for tracking the target vehicle based on color and the scale-invariant feature transform (SIFT), (2) a control system based on linear quadratic Gaussian control, and (3) a path following system. Their experiments showed that the mean and maximum error in the visual distance estimate were 0.72 m and 2.42 m, respectively, the follower speed was between 7.6 km/h and 10.2 km/h, and the follower separation was between 10.46 m and 23.71 m.

Taken together, although some research has been carried out on vehicle localization, more detailed empirical investigations are needed to dynamically estimate the target vehicle's geolocation in a GPS coordinate system via a monocular camera with the purpose of generating data to model the traffic scene and improve the ITMS performance.

\section{Research strategy and methodology}

In this paper, we proposed two new approaches by integrating deep learning, image processing, and geometric computation to use the vision sensing and self-localization capabilities of a mobile ego vehicle to estimate the geolocation of target vehicles. Figure \ref{research strategy} illustrates our proposed research strategy. The components included in Figure \ref{research strategy} are as follows:

\begin{figure*}[hbt!]
\centering
  \includegraphics[width=\textwidth]{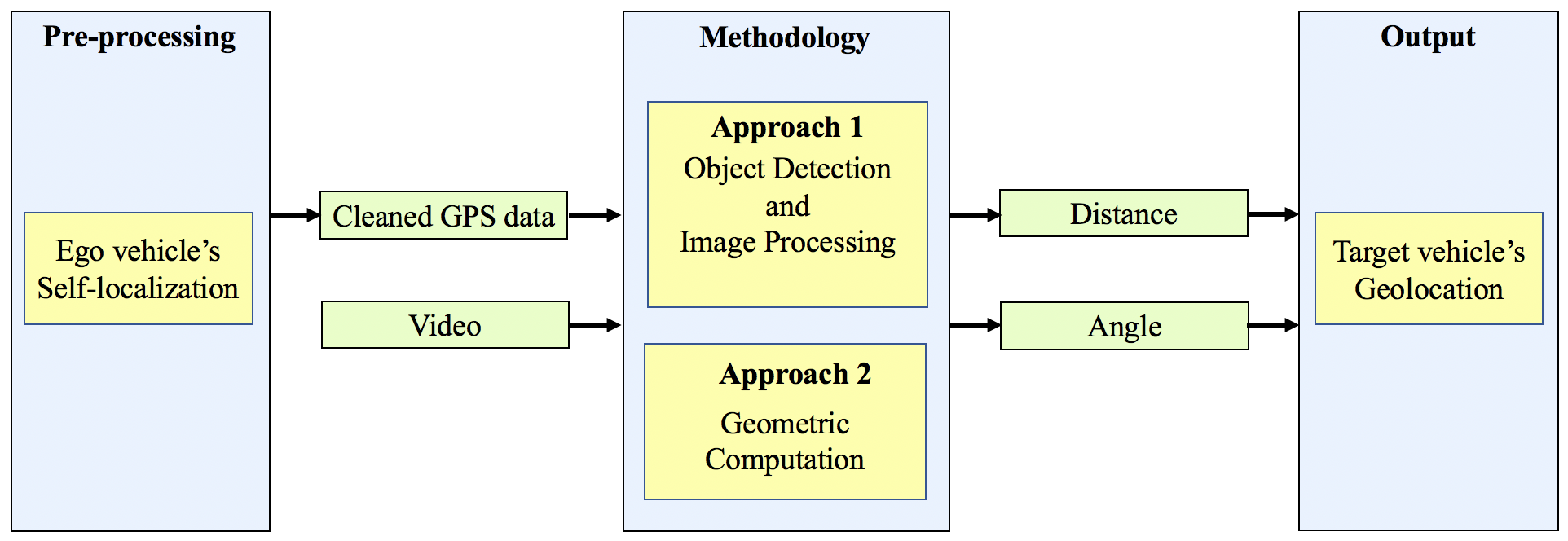}
  \caption{The proposed steps in our research strategy.}
  \label{research strategy}
\end{figure*}

\subsection{Pre-processing}
For the estimation of the target vehicle's geolocation, the geolocation of the ego vehicle is required. The collected latitude and longitude of the ego vehicle, which are usually collected by a GPS receiver, might be noisy. Collecting the geolocations of the ego vehicle accurately plays a vital role in accurately estimating the target vehicle's geolocations. Therefore, to enhance the accuracy of the ego-vehicle localization in the GPS coordinate system, the proposed approach in \cite{namazi2021gps}, based on cross-GPS validation, interpolation, best-fit, and map-matching techniques, is used. 

\subsection{Methodology}

\begin{figure}[hbt!]
    \centering
    \includegraphics[scale=0.3]{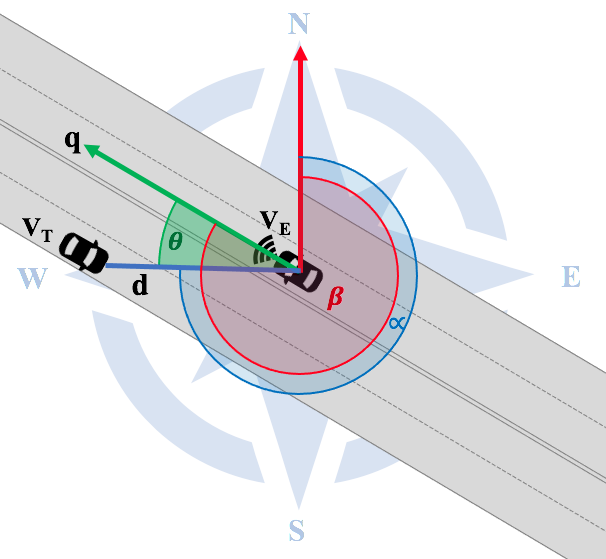}
    \caption{The required parameters for estimating the target vehicle's geolocation.}
    \label{angles}
\end{figure}

As shown in Figure \ref{angles}, to estimate the target vehicle's geolocation in the GPS coordinate system, in addition to the ego vehicle's geolocation, the distance \(d\) between the ego vehicle \(V_E\) and the target vehicle \(V_T\) and the clockwise angle \(\alpha\) between the north (N) and \(d\) are required \cite{distance}.

In this regard, as Figure \ref{research strategy} shows, we proposed two approaches, as follows:

\begin{enumerate}[label=\Roman*)]
    \item {Approach 1: Object detection and image processing}

    \begin{itemize}
         \item Estimating the distance \(d\)\\
    To begin the distance estimation process, we employed you only look once (YOLO)-v3 \cite{redmon2016you}\cite{redmon2018yolov3} to detect target vehicles via the ego vehicle's vision. YOLO-v3 is a well-documented open-source one-stage method for detecting objects. \say{YOLO-v3 is extremely fast and accurate} \cite{yoloweb}. For example it is more than 1000x faster than R-CNN and 100x faster than Fast R-CNN \cite{redmon2016you}\cite{redmon2018yolov3}. Wang et al., \cite{wang2021daedalus} listed YOLO-v3 as the second most popular object detector models. Therefore, we chose to use YOLO-v3 in this study.
  
     We trained YOLO-v3 on the KITTI dataset \cite{kitti}, as this study focuses on traffic objects, and KITTI includes eight categories of traffic objects: car, van, truck, pedestrian, person\_sitting, cyclist, tram, and misc \cite{kitti}. To estimate the distance \(d\) from the ego vehicle to the target vehicle, we followed the approach proposed by Namazi et al. \cite{namazi2019master}. This approach \cite{namazi2019master} was based on the pinhole camera model by considering the real, pre-known size of the target vehicle and the size of the bounding box added by YOLO-v3 around the target vehicle on the image plane. Distance \(d\) was calculated based on the average of the computed distances for both vehicle width and vehicle height by using a weight factor (i.e., 85\% of the height and 15\% of the width).

    \item {Estimating the angle \(\alpha\)}
        
        As presented in Figure \ref{angles}, in order to estimate the clockwise angle \(\alpha\) between the north N and \(d\), we need to estimate angle \(\beta\), which is the angle between the north N and the ego vehicle \(V_E\)'s movement direction q, as well as angle \(\theta\), which is the angle between \(d\) and q. 
        
    \begin{enumerate}[label=\roman*.]
        \item {Estimating the angle \(\beta\)}
        
        To estimate the angle \(\beta\), we need to identify the movement direction of the ego vehicle \(V_E\) based on its collected GPS coordinates in sequential frames as a start point (\(\phi_1,\lambda_1\)) and an end-point (\(\phi_2,\lambda_2\)) for all frames. We used Eq. \ref{M} - Eq. \ref{e_beta} \cite{distance} to estimate angle \(\beta\) along the whole trajectory dynamically.
        \begin{equation}
            M=\sin{(\lambda_2 - \lambda_1)}\cdot\cos\phi_2
            \label{M}
        \end{equation}
        \begin{equation}
            N=\cos{\phi_1}\cdot\sin{\phi_2}-\sin{\phi_1}\cdot\cos{\phi_2}\cdot\cos{(\lambda_2 - \lambda_1)}
            \label{N}
        \end{equation}
        \begin{equation}
        \begin{split}
            \beta = \atantwo(M,N)
        \end{split}
        \label{e_beta}
        \end{equation}
        \item {Estimating the angle \(\theta\)}
        
        The idea of estimating angle \(\theta\) in our first approach is presented in Figure \ref{theta}. In Figure \ref{theta}, the blue bounding box shows the target vehicle \(V_T\). P is the central point on the button edge of the bounding box around the target vehicle \(V_T\), and \(H\) is the central point of the image. Angle \(\theta\) is estimated based on the horizontal angle per pixel (\(\gamma\)) in degrees and on the number of horizontal pixels between \(P\) and the vertical line passing through \(H\), shown by a red line \(T\). \(\gamma\) is estimated based on the camera's horizontal field of view (FOV) and the video's resolution. In this study, the video's resolution was \(960\times720\) pixels, and the camera's horizontal FOV was 86.7 degrees \cite{FOV}. Therefore, \(\gamma\) is equal to 0.90 degrees. Angle \(\theta\) in degrees is estimated as follows:
        \begin{equation}
            \theta = T \cdot \gamma
        \end{equation}

        \begin{figure}[hbt!]
            \centering
            \includegraphics[scale=0.5]{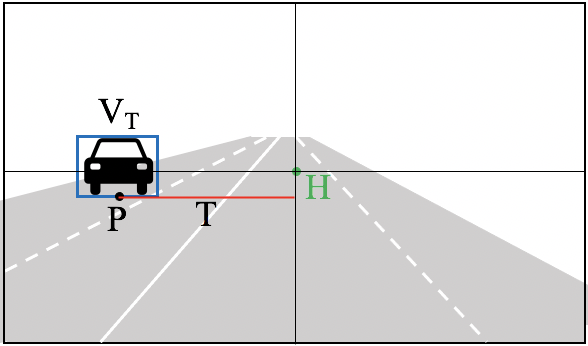}
            \caption{The parameters used to estimate angle \(\theta\) between the ego vehicle and the target vehicle used in Approach 1.}
            \label{theta}
        \end{figure}

        \item {Estimating the angle \(\alpha\)}
        
        To estimate the angle \(\alpha\), we considered three different conditions, as follows:
        \begin{itemize}
            \item If the target vehicle drives in the same lane as the ego-vehicle, then \(\alpha=\beta\) and \(\theta=0\).
            \item 	If the target vehicle drives on the left side of the ego vehicle, then, as Figure \ref{pic_alpha}, (a) shows, \(\alpha=\beta-\theta\).
            \item If the target vehicle drives on the right side of the ego vehicle, then, as Figure \ref{pic_alpha}, (b) shows,  \(\alpha=\beta+\theta\).
        \end{itemize}
        
        \begin{figure*}[hbt!]
            \centering
            \includegraphics[width=\textwidth]{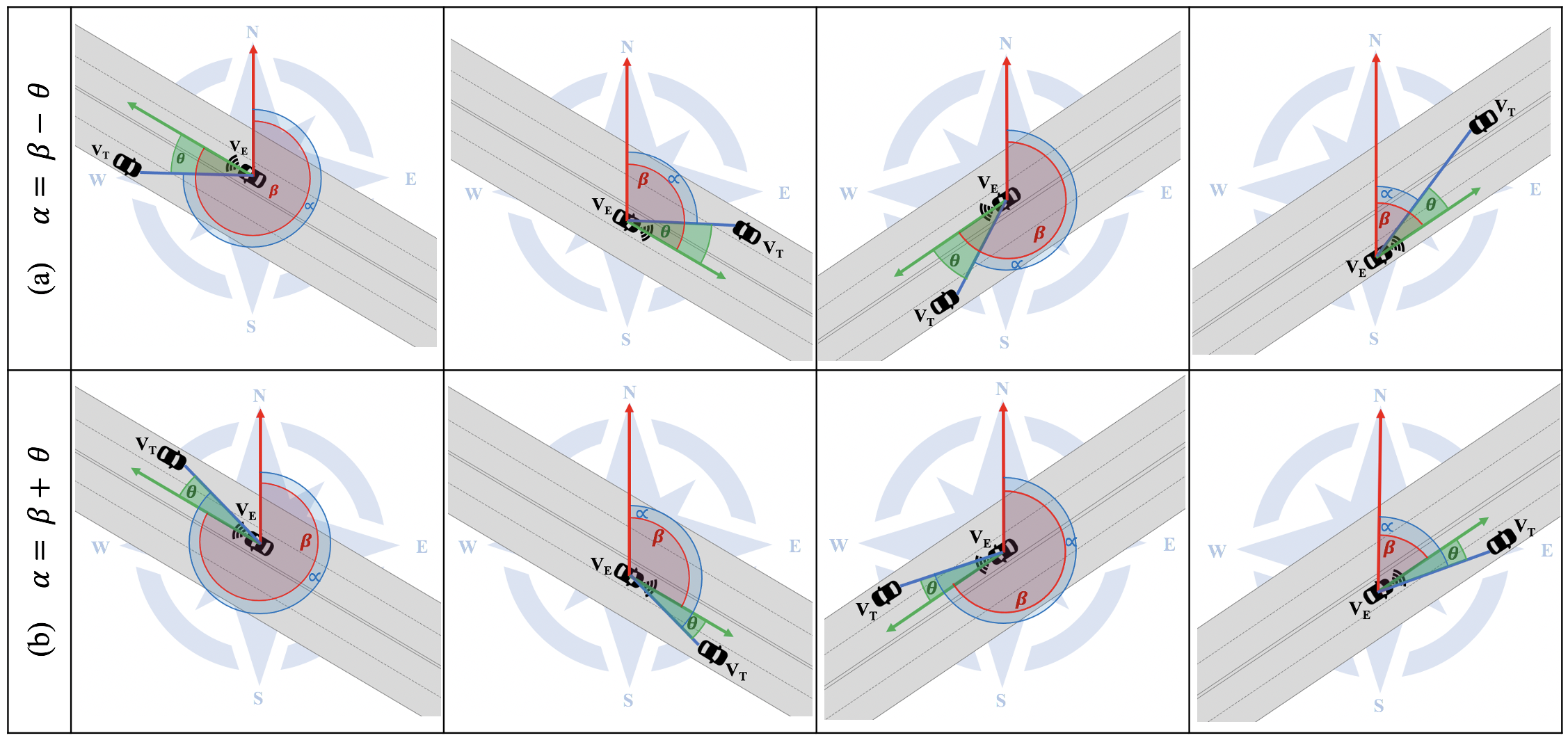}
            \caption{The mathematical relations between \(\alpha\), \(\beta\), and \(\theta\).}
            \label{pic_alpha}
        \end{figure*}
    \end{enumerate}
    \end{itemize}

    \item {Approach 2: Geometric computation}
    
    The main idea of this approach is to transform 2D
pixel coordinates of point P into 3D world coordinates. By assessing the 3D world coordinates of point P, we would be able to estimate distance \(d\) and angle \(\alpha\), which are needed to estimate the target vehicle's geolocation. 

In this regard, we utilized a pinhole camera model as shown in Figure \ref{pinhole}. In this figure, C is the perspective center of the camera and the origin of the camera coordinate frame (CCF). Three unit vectors
of the CCF are represented by \(y_1\), \(y_2\), and \(y_3\). The image coordinate frame (ICF) is centered at principal point H with unit vectors \(r_1\) and \(r_2\). The principal axis passes through C and H and is perpendicular to the image plane. The distance from C to the image plane is \(f\), which is the camera's focal length.
The image plane carries a 2D pixel coordinate frame (PCF) with unit vectors \(z_1\) and \(z_2\). The image plane is subdivided into \(n_h\) pixels horizontally and \(n_v\) pixels vertically. To project the detected vehicle on the image onto the real world, we need to transform point P with pixel coordinates \((p_1,p_2)\) (which is the central point on the button edge of the bounding box around the target vehicle) into a 3D world coordinate
representation \((w_1,w_2,w_3)\).

\begin{figure}[hbt!]
    \centering
    \includegraphics[scale=0.4]{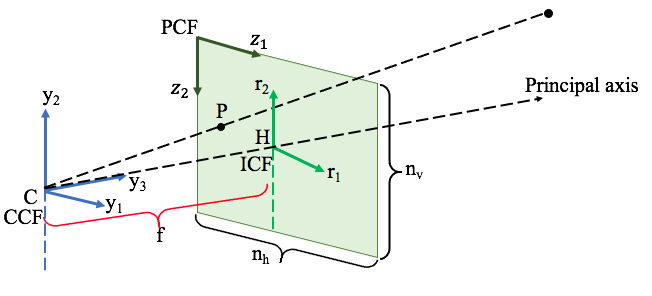}
    \caption{The pinhole camera model used in Approach 2.}
    \label{pinhole}
\end{figure}

In this regard, we first need to identify the 3D coordinates of point P in 3D camera coordinates. Based on Figure \ref{pcf2icf}, the 3D coordinates of point P in 3D camera coordinates are presented in Eq. \ref{camcoor}.

\begin{equation}
    \begin{pmatrix}
    y_1\\
    y_2\\
    y_3
    \end{pmatrix}=\begin{pmatrix}
    p_1 - h_1\\
    -(p_2 - h_2)\\
    f
    \end{pmatrix}
    \label{camcoor}
\end{equation}

\begin{figure}[hbt!]
    \centering
    \includegraphics [scale=0.4]{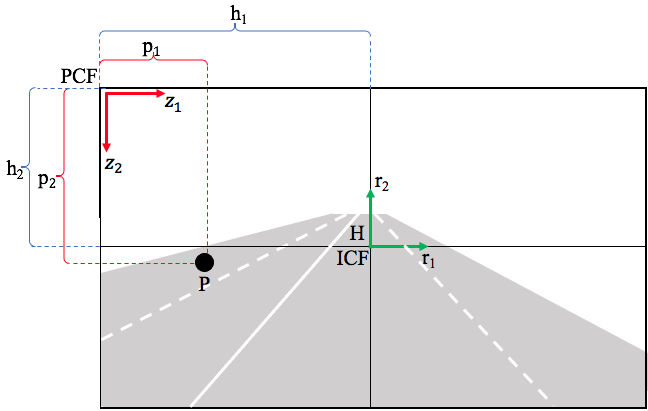}
    \caption{The image plane used in Approach 2.}
    \label{pcf2icf}
\end{figure}

In the follow-up step, we need to identify the 3D coordinates of point P in 3D world coordinates. In this step, we temporally assumed that the camera's pitch angle, yaw angle, and roll angle were equal to 0. The height of the camera mounted on the ego vehicle from the road surface is named \(h\), and we assumed that the world coordinates are located on the road surface. The 3D coordinates of point P in 3D world coordinates (\(w_1,w_2,w_3\)) are given in Eq. \ref{pto3D}.

\begin{equation}
   \begin{pmatrix}
   w_1\\
   w_2\\
   w_3
   \end{pmatrix}=\begin{pmatrix}
    p_1 - h_1\\
    -(p_2 - h_2)+h\\
    f
    \end{pmatrix}
    \label{pto3D}
\end{equation}

After that, we need to find a mathematical expression for all points that lie on the viewing ray from camera center C through point P in world coordinates, as presented in Eq. \ref{mathexpresion1}, where \(\xi\) defines any position along the viewing ray.
\begin{equation}
 \begin{pmatrix}
   w_1\\
   w_2\\
  w_3\end{pmatrix}= \begin{pmatrix}
   0\\
   h\\
  0
   \end{pmatrix} +\xi\cdot\begin{pmatrix}
    p_1 - h_1\\
    -(p_2 - h_2)\\
    f
    \end{pmatrix}   
    \label{mathexpresion1}
\end{equation}

As we assumed, the world coordinate system is located on the road surface; therefore, the height of any point on the road surface in the world coordinate system is equal to 0. So we can express Eq. \ref{mathexpresion1} as  Eq. \ref{mathexpresion2}.
\begin{equation}
\begin{pmatrix}
   w_1\\
   0\\
  w_3\end{pmatrix}=
    \begin{pmatrix}
   0\\
   h\\
  0
   \end{pmatrix} +\xi\cdot\begin{pmatrix}
    p_1 - h_1\\
    -(p_2 - h_2)\\
    f
    \end{pmatrix} 
    \label{mathexpresion2}
\end{equation}

Finally, to increase the accuracy of this estimation, we need to consider the camera's pitch angle \(\sigma\).  Rotating the camera by angle \(\sigma\) has no effect on the location of point P in the \(w_1\) direction, but the point's location in the \(w_2\) and \(w_3\) directions will be affected by this rotation. Therefore, the following rotation is applied:
\begin{equation}
\begin{pmatrix}
   w_1\\
   0\\
  w_3\end{pmatrix}=
    \begin{pmatrix}
   0\\
   h\\
  0
   \end{pmatrix} +\xi\cdot\begin{pmatrix}
   1 && 0 && 0\\
   0 && \cos(-\sigma) && -\sin(-\sigma)\\
   0 && \sin(-\sigma) && \cos(-\sigma)
   \end{pmatrix}
   \cdot \begin{pmatrix}
    p_1 - h_1\\
    -(p_2 - h_2)\\
    f
    \end{pmatrix}  
\end{equation}
Therefore, \(\xi\), \(w_1\), and \(w_3\) are calculated as follows:
\begin{equation}
    \xi = \frac{-h}{(\cos(-\sigma)\cdot(-p_2 + h_2)-\sin(-\sigma)\cdot f)}
\end{equation}
\begin{equation}
    w_1=\xi\cdot (p_1 - h_1)
\end{equation}
\begin{equation}
    w_3= \xi\cdot (\sin(-\sigma)\cdot (-p_2 + h_2) + \cos(-\sigma)\cdot f)
\end{equation}

To calculate \(\xi\), \(w_1\), and \(w_3\), we need to estimate the camera's height from the road surface \(h\), the camera's pitch angle \(\sigma\), and the camera's focal length \(f\). 
\begin{itemize}
    \item Estimating the camera's focal length \(f\)
    
    In Approach 2, the camera's focal length \(f\) in pixels is calculated based on the trigonometric relation presented in Eq. (\ref{eq.focal}). We used the horizontal number of pixels \(n_h\) from the video's resolution and the camera's horizontal FOV \(\rho\) in degrees \cite{FOV}. 
    \begin{equation}
   f =\frac{n_h}{2\cdot \tan({\frac{\rho}{2}})}
   \label{eq.focal}
\end{equation} 

\item Estimating the camera's pitch angle \(\sigma\)

    To estimate the camera's pitch angle \(\sigma\), we used a vanishing point estimated based on the lane detection.

      To detect lanes, as we presented in \cite{namazi2019master} and \cite{namazi2020lane}, we used canny edge detection \cite{ding2001canny} and the progressive probabilistic Hough transform \cite{galamhos1999progressive}\cite{matas2000robust}. In this study, we go further to identify the vanishing point based on the detected parallel lines on the road nearby the ego vehicle.  
    
    To estimate the pitch angle \(\sigma\), we used the camera's focal length (f) and the vertical differences between the principal point \(H = (h_1, h_2)\) and the vanishing point \(J = (j_1, j_2)\), as shown in Figure \ref{pitch}. In this figure, the blue lines represent the detected parallel lines on the road nearby the ego vehicle in a perspective view. Based on this figure, the camera's pitch angle \(\sigma\) can be calculated by Eq. \ref{pitcheq}. 

\begin{equation}
    \sigma = \atantwo({j_2 - h_2},{f})
    \label{pitcheq}
\end{equation}

  \begin{figure}[hbt!]
    \centering
    \includegraphics[scale=0.35]{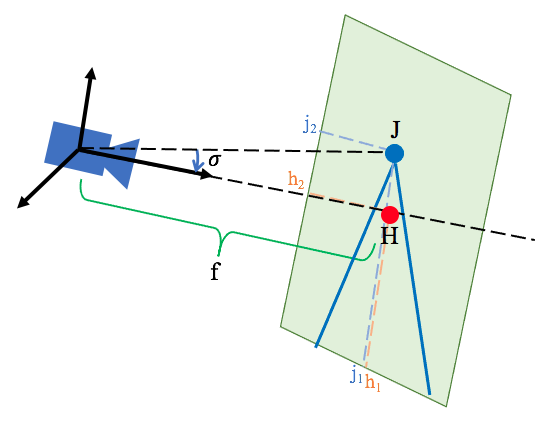}
    \caption{The camera's pitch angle \(\sigma\) and vanishing point used in Approach 2.}
    \label{pitch}
\end{figure}

    \item Estimating the camera's height \(h\) from the road surface 

    To estimate the height \(h\) of the camera mounted on the ego vehicle from the road surface by considering the camera's pitch angle \(\sigma\), we applied Thales's theorem \cite{thalestheorem}. Thales's theorem in this context is presented in Figure \ref{distancepitch}. The variables in this figure are defined as follows:
    
\begin{equation}
    A=\frac{f}{\cos\sigma}
\end{equation}
\begin{equation}
    B=h\cdot\tan\sigma
\end{equation}
\begin{equation}
    E=\frac{h}{\cos\sigma}
\end{equation}
\begin{equation}
    G=f\cdot \tan\sigma
\end{equation}
\begin{equation}
    K=G + N
\end{equation}

\begin{figure*}
            \centering
            \includegraphics[width=\textwidth]{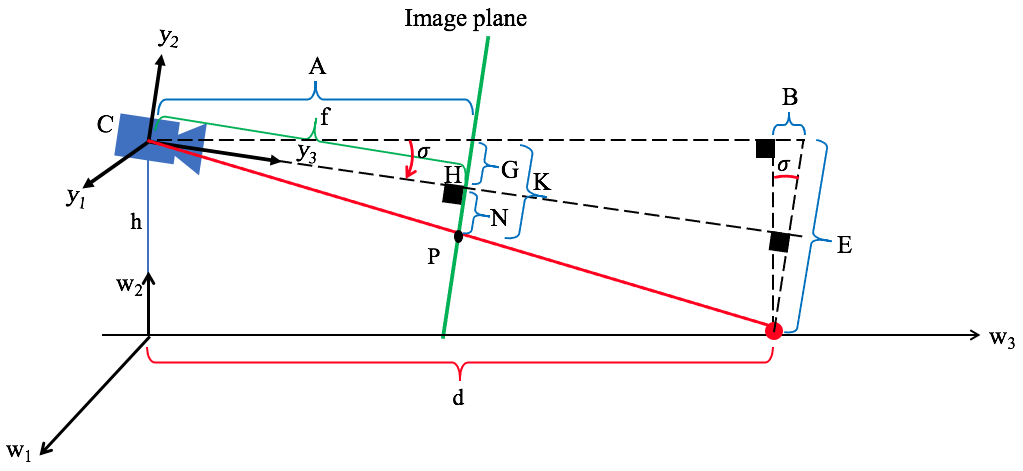}
            \caption{The camera's height from the road surface by considering the camera's pitch angle \(\sigma\) used in Approach 2.}
            \label{distancepitch}
        \end{figure*}
To estimate the camera's height \(h\) based on Thales's theorem and the estimated distance \(d\) by Approach 1, we have the following equation.

\begin{equation}
    \frac{K}{E}=\frac{A}{d+B}
    \label{thales}
\end{equation}

By simplifying Eq. \ref{thales}, \(h\) in meters is calculated as follows:
\begin{equation}
h= \frac{(N+f\cdot \tan \sigma) \cdot d  \cdot {(\cos\sigma)^2}}{f - (N+f\cdot \tan \sigma) \cdot\tan\sigma \cdot {(\cos\sigma)^2}}
\end{equation}

\end{itemize}
\end{enumerate}

Finally, by estimating the camera's height from the road surface \(h\), the camera's pitch angle \(\sigma\), and the camera's focal length \(f\), we can calculate \(\xi\), \(w_1\), and \(w_3\). Because \(w_1\) and \(w_3\) represent point P in 3D world coordinates, where \(w_2\)=0, we can estimate distance \(d\) by Approach 2 based on the Euclidean distance between \(w_1\) and \(w_3\) as presented in Eq. \ref{dapp2} \cite{Euclideandis}, and estimate angle \(\theta\) based on trigonometry  presented in Eq. \ref{thetaapp2} \cite{Trigonometry}. Finally, angle \(\alpha\) can be estimated based on the proposed conditions in Section 3.2.I.iii.

\begin{equation}
             d  = \sqrt {w_1^2 + w_3^2} 
             \label{dapp2}
        \end{equation}

        \begin{equation}
            \theta = \atantwo{({w_3},{w_1})}
            \label{thetaapp2}
        \end{equation}

\subsection{Estimating the geolocation of the target vehicle}
To estimate the target vehicle's geolocation with both approaches, we used Eq.  \ref{eq3} - Eq. \ref{eq4} \cite{distance}. In these formulas, the variables are as below:

(\(\ell_1, g_1\)) represent the geolocation of the ego vehicle 

(\(\ell_2, g_2\)) represent the geolocation of the target vehicle 

\(R\) represents the Earth's radius

\(d\) represents the estimated distance between the ego vehicle and the target vehicle by both approaches

\(\alpha\) represents the estimated angle between the north N and \(d\) by both approaches

\begin{equation}
    \ell_2=\asin(\sin(\ell_1)\cdot\cos(d/R) + \cos(\ell_1)\cdot\sin(d/R)\cdot\cos(\alpha))
    \label{eq3}
\end{equation}

\begin{equation}
    U=\sin(\alpha)\cdot\sin(d/R)\cdot\cos(\ell_1)
    \label{B}
\end{equation}
\begin{equation}
    V= \cos(d/R)-\sin(\ell_1)\cdot\sin(\ell_2)
    \label{A}
\end{equation}
\begin{equation}
    g_2=g_1 + \atantwo(U, V)
    \label{eq4}
\end{equation}

\section{Experiments and results}

In this paper, we carried out experiments using real-world traffic data to demonstrate the effectiveness of both proposed approaches for estimating the target vehicle's geolocation by an ego vehicle's vision. 

\subsection{Data collection}
We used three vehicles and drove them by the following the pre-defined scenarios in Chengdu, China. All vehicles were equipped with two GoPro Hero 7 cameras, and each camera included a built-in GPS receiver. We used the GoPro Hero 7 camera as it provides us with both visual information and GPS data of the vehicles. One of the cameras mounted on the front window glass looked forward through the window, and another mounted on the back window glass looked backward. The purpose of mounting two cameras on the same vehicle was to improve the ego vehicle's self-location in the pre-processing step \cite{namazi2021gps}, and to collect more data for future studies. To estimate the target's geolocation, this study used only the footage collected from the camera mounted on the front window glass. We used one of these vehicles as an ego vehicle. The other two vehicles were treated as target vehicles. The GPS data collected by the target vehicles were used as ground truth to assess our proposed approaches.

The settings of the GoPro Hero 7 cameras we used were as follows. The used mode was \(4 \times 3\) and linear with a zoom = 0\%. During the recording of the footage, the video's resolution and the frame rate were \(1920\times1440\) and 60 frames per second (FPS), respectively. 
We adjusted the video's resolution to \(960\times720\) and the frame rate to 1 FPS to apply pre-processing and vehicle detection.

\begin{figure}[ht!]
\centering
\includegraphics[scale=0.25]{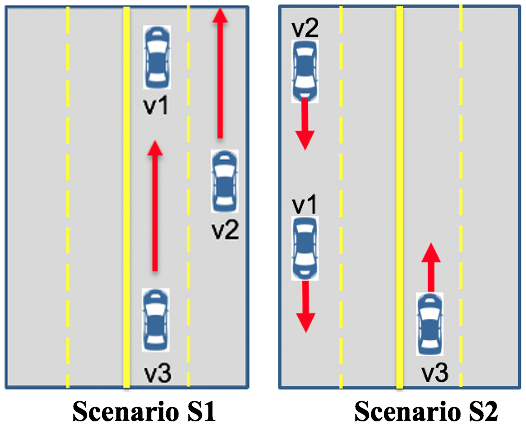}
\caption{The scenarios studied using both approaches.}
\label{S4}
\end{figure}

Figure \ref{S4} shows the studied scenarios, called Scenario S1 and Scenario S2.  In Scenario S1, ego vehicle v3 and target vehicles v1 and v2 are driven in the same direction on a straight trajectory. The purpose of this scenario was to evaluate our proposed approaches with one of the target vehicles driving on the same lane as the ego vehicle and the other target vehicle driving on the next lane. In Scenario S2, ego vehicle v3 and target vehicles v1 and v2 are driven in opposite directions on a straight trajectory. In the scenario in which the vehicles are driven in opposite directions, the period between detecting a target vehicle via an ego vehicle until both vehicles pass each other is short, so the number of estimated locations is limited to this short period. 

\subsection{Experiments}
The experiments were run using a laptop with a 3.1 GHz Intel Core i5 processor and Intel Iris Plus Graphics 650 1536 MB. As we explained in Section 4.1, we chose to use 1 FPS when analyzing the video, with the purpose of making a trade-off between the amount of generated data and the running time of the system. Also, as the vehicles' speeds were low (between 19.55 km/h and 30.18 km/h on average), there were hardly any informative changes in the vehicle's speed, distance, angle, and location within less than one second. Therefore, analyzing the data with a higher frequency could not provide much extra information. We measured the running time based on the experimental studies using Approaches 1 and 2 after the data pre-processing. We found that the system based on Approach 1 ran 0.680 FPS on average, and that the system based on Approach 2 ran 0.684 FPS on average, to output the geolocation of the target vehicle from the input videos and the pre-processed ego vehicle's geolocations. 

\subsubsection{Evaluation}
The evaluation is done in two steps: (1) plotting the estimated geolocations (i.e., latitude and longitude) of the target vehicles on Google Maps and (2) analyzing the distance vector between the estimated target vehicle's geolocations by both proposed approaches and the ground-truth data. 

\begin{figure}[ht!]
\centering
\includegraphics[width=\linewidth]{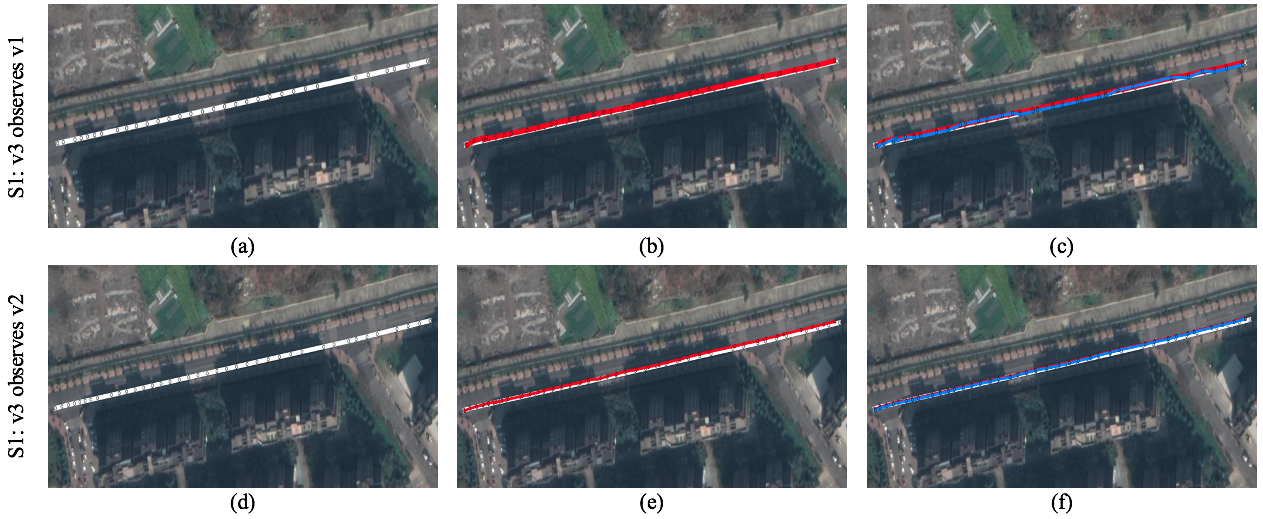}
\caption{The estimated trajectory with Scenario S1 on the map. Two cases are considered: (1) ego vehicle v3 observes target vehicle v1 (a-c) and (2) ego vehicle v3 observes target vehicle v2 (d-f).}
\label{outputsmaps1}
\end{figure}

\begin{figure}[ht!]
\centering
\includegraphics[width=\linewidth]{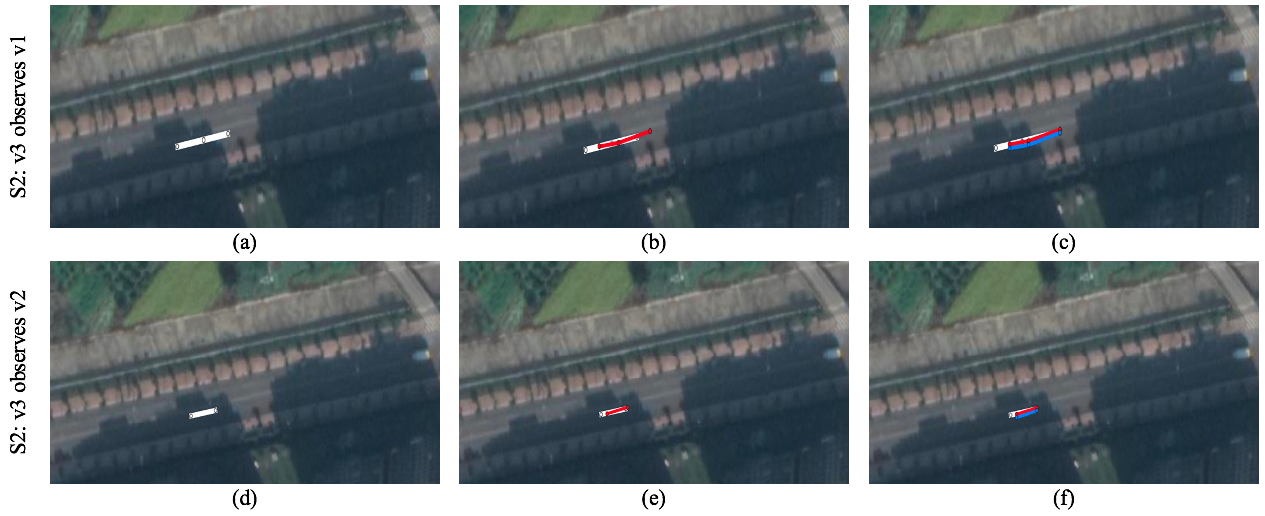}
\caption{The estimated trajectory with Scenario S2 on the map. Two cases are considered: (1) ego vehicle v3 observes target vehicle v1 (a-c) and (2) ego vehicle v3 observes target vehicle v2 (d-f).}
\label{outputsmaps2}
\end{figure}

Figure \ref{outputsmaps1} and Figure \ref{outputsmaps2}  present the outputs of the experiments related to Scenario S1 and Scenario S2, respectively. In these figures, the white polyline in (a) and (d) shows the ground-truth trajectory of the observed target vehicle. The red polyline in (b) and (e) shows the trajectory of the target vehicle estimated with Approach 1. The blue polyline in (c) and (f) shows the trajectory of the target vehicle estimated with Approach 2. These figures show that the  trajectories of the target vehicle estimated with both approaches are plotted on the correct lane of the road and that they almost overlapped with the ground-truth trajectory. This means that both approaches enable us to estimate the trajectory of the target vehicle accurately on the right lane of the road. As expected, with Scenario 2, the number of plotted positions along the trajectory, presented in Figure \ref{outputsmaps2}, are limited (2-3 points) because of the opposite movement directions of the ego vehicle and target vehicle and the short sensing time.

\begin{figure}[ht!]
\centering
\includegraphics[width=\linewidth]{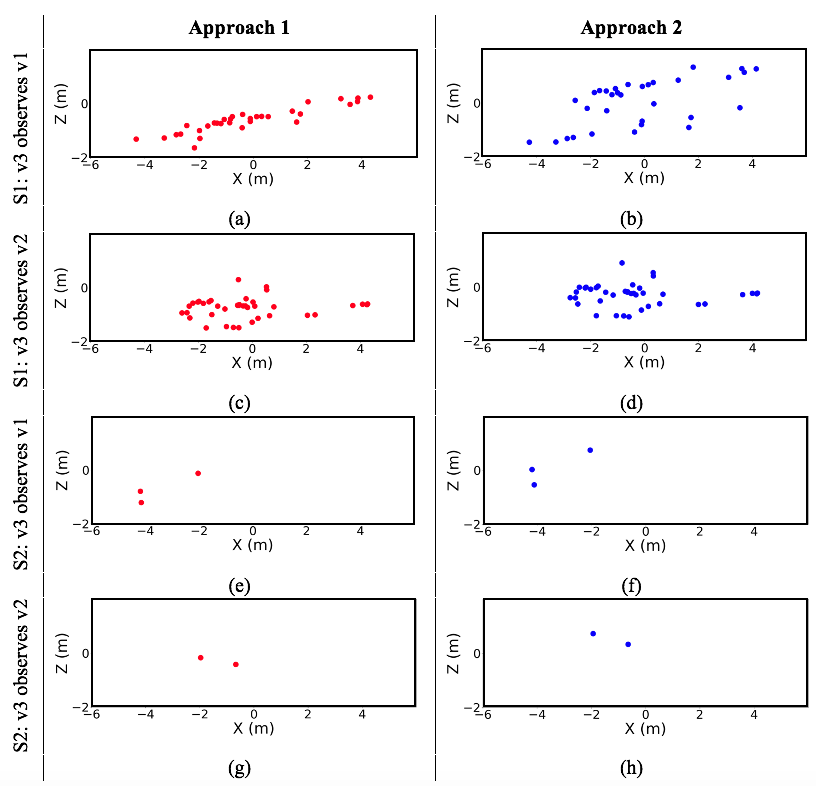}
\caption{The distance vectors between ground truth and
estimated geolocations with Approach 1 (a, c, e, and g) and Approach 2 (b, d, f, and h) for Scenario 1 and Scenario 2. Two cases are considered: (1) ego-vehicle v3 observes target vehicle v1, and (2) ego-vehicle v3 observes target vehicle v2. X shows longitudinal and Z shows lateral directions
}
\label{outputsplt}
\end{figure}

To analyze the estimated geolocations of the target vehicle numerically, we used the distance vector between the ground truth and the geolocations estimated by both approaches. This analysis provides information regarding the estimation deviation in our proposed approaches. Figure \ref{outputsplt} visualizes our findings related to Scenario S1 and Scenario S2. Our numerical findings are summarized in Table \ref{rmse}, as well. As Figure \ref{outputsplt} and Table \ref{rmse} show, the geolocation estimation deviation (based on the absolute values) with Approach 1 is on average between 1.38 m and 3.54 m. The geolocation estimation deviation with Approach 2 is on average between 1.4 m and 3.51 m. Figure \ref{outputsplt} (a) and (b) shows a slightly upward trend between the plotted points. This result may be explained by the fact that the collected data by a GPS receiver to provide the ego vehicle's location and ground truth data of the target vehicle's position were not noise-free. As we expected, Figure \ref{outputsplt} (e) - (h) represent the limited points as Scenario S2 focused on studying the vehicle movement in the opposite directions and the sensing lifetime was limited.  In addition, as Table \ref{rmse} shows, the highest on average geolocation estimation deviation with both approaches is obtained in Scenario S2, when v3 observes v1. A possible explanation for this might be that as in this scenario, only limited geolocations (2-3) were estimated, so the estimation deviation of one point has a big effect on the average error.

\begin{table}[]
\centering
\caption{Evaluation results.}
\resizebox{\textwidth}{!}{%
\begin{tabular}{cccccccccc}
\hline
\textbf{S\#} & \textbf{V\#} & \multicolumn{4}{c}{\textbf{Estimation deviation of Approach 1 (m)}} & \multicolumn{4}{c}{\textbf{Estimation deviation of Approach 2 (m)}} \\ \cline{3-10} 
\textbf{}    & \textbf{}    & \textbf{Min}   & \textbf{Avg}  & \textbf{Max}  & \textbf{RMSE}  & \textbf{Min}   & \textbf{Avg}   & \textbf{Max}   & \textbf{RMSE}   \\ \hline
\multirow{2}{*}{\textbf{S1}} & \textbf{v3 observes v1} & 0.50  & 2.03 & 4.51 & 2.35 & 0.35 & 2.02 & 4.51 & 2.34 \\
                             & \textbf{v3 observes v2} & 0.47  & 1.74 & 4.31 & 2.03 & 0.19 & 1.63 & 4.18 & 1.96 \\
\multirow{2}{*}{\textbf{S2}} & \textbf{v3 observes v1} & 2.040 & 3.54 & 4.33 & 3.70 & 2.18 & 3.51 & 4.20 & 3.63 \\
                             & \textbf{v3 observes v2} & 0.79  & 1.38 & 1.97 & 1.50 & 0.72 & 1.4  & 2.07 & 1.55 \\ \hline
\end{tabular}%
}
\label{rmse}
\end{table}

 To analyze our proposed approaches further, we applied the root mean square error (RMSE) to the distance vector between the estimated geolocations and the ground truth to show the estimation deviation. The calculated RMSE related to Approach 1 was between 1.5 m and 3.7 m (2.39 m on average). The calculated RMSE related to Approach 2 was between 1.55 m and 3.63 m (2.37 m on average). Overall, these results indicate that, in the studied scenarios, Approach 2 is slightly (about 0.02 m on average) better than Approach 1.

\begin{figure}[ht!]
\centering
\includegraphics[width=\linewidth]{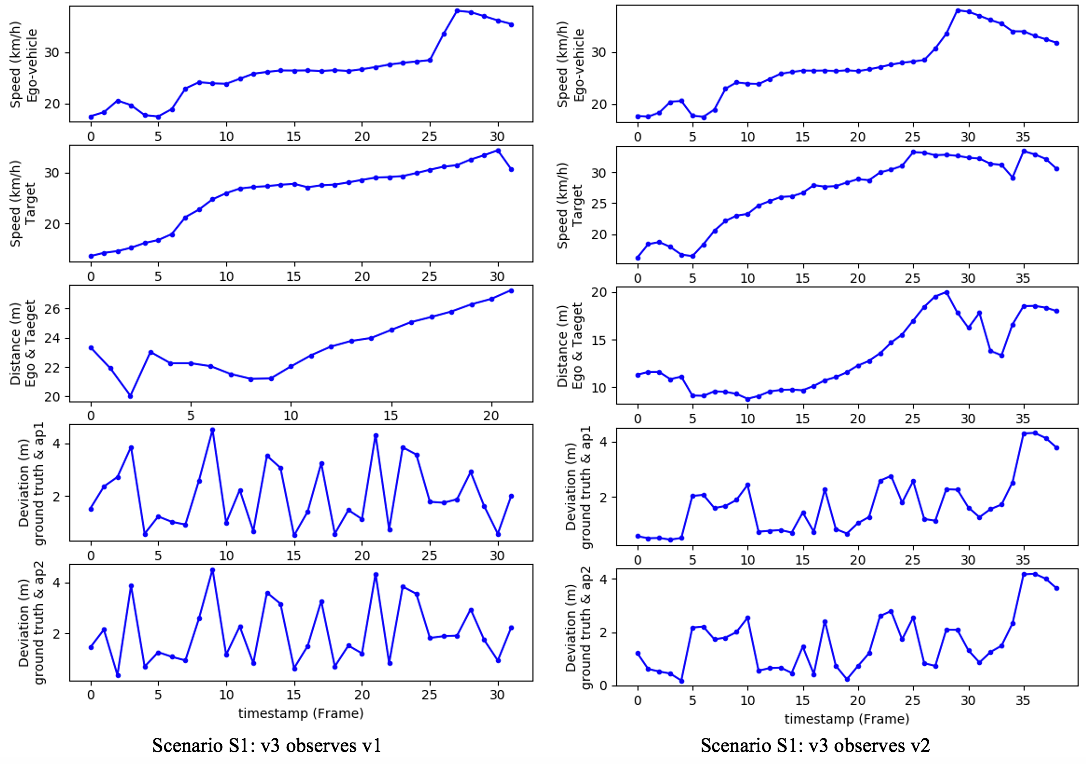}
\caption{Plotting the deviations with regard to ground truth to study the effect of speed and distance  (here shown for Scenario S1.)}
\label{speeddiseffectstd}
\end{figure}

As the speed of the vehicles and the distance between them may influence our estimation accuracy, we studied the effect of the speed of the ego vehicle and target vehicle and the distance between them on estimating the target vehicle's geolocation. As an example, we presented our findings related to Scenario 1 in Figure \ref{speeddiseffectstd}. Figure \ref{speeddiseffectstd} shows that in the case in which v3 observes v1, with both vehicles driving on the same lane, changes in the distance and speed have no significant effect on our estimation accuracy. However, when v3 observes v2, with both vehicles driving on different lanes, increasing the distance between the vehicles, as caused by changes in the vehicles' speed, increases the estimation deviation. However, as the speed of the vehicles and the distance between them were limited in the studied scenarios, more studies are needed to validate these findings in the future. The extracted traffic data (i.e., the ego vehicle's and the target vehicle's speed and the distance between the vehicles) are summarized in Table \ref{speed&dis}.

\begin{table}[]
\centering
\caption{Traffic data measurements with Scenario 1 and Scenario 2.}
\resizebox{\textwidth}{!}{%
\begin{tabular}{ccccccccccc}
\hline
\textbf{S\#} &
  \textbf{V\#} &
  \multicolumn{3}{c}{\textbf{Ego vehicle speed (km/h)}} &
  \multicolumn{3}{c}{\textbf{Target vehicle speed (km/h)}} &
  \multicolumn{3}{c}{\textbf{Dist. ego and target vehicles (m)}} \\ \cline{3-11} 
\textbf{} &
  \textbf{} &
  \textbf{Min} &
  \textbf{Avg} &
  \textbf{Max} &
  \textbf{Min} &
  \textbf{Avg} &
  \textbf{Max} &
  \textbf{Min} &
  \textbf{Avg} &
  \textbf{Max} \\ \hline
\multirow{2}{*}{\textbf{S1}} & \textbf{v3, v1}  & 17.50 & 26.38 & 38.04 & 13.64 & 25.65 & 34.35 & 20.04 & 25.02 & 38.84 \\
                             & \textbf{v3 , v2} & 17.50 & 27.14 & 38.04 & 16.19 & 26.94 & 33.4  & 8.81  & 13.25 & 20.01 \\
\multirow{2}{*}{\textbf{S2}} & \textbf{v3, v1}  & 16.98 & 19.55 & 22.04 & 29.12 & 29.40 & 29.55 & 17.35 & 30.78 & 44.14 \\
                             & \textbf{v3 , v2} & 22.04 & 23.2  & 24.36 & 29.63 & 30.18 & 30.74 & 20.53 & 27.95 & 35.37 \\ \hline
\end{tabular}%
}
\label{speed&dis}
\end{table}

\subsubsection{Comparison of Approach 1 with Approach 2}

\begin{figure}[hbt!]
\centering
  \includegraphics[width=\textwidth]{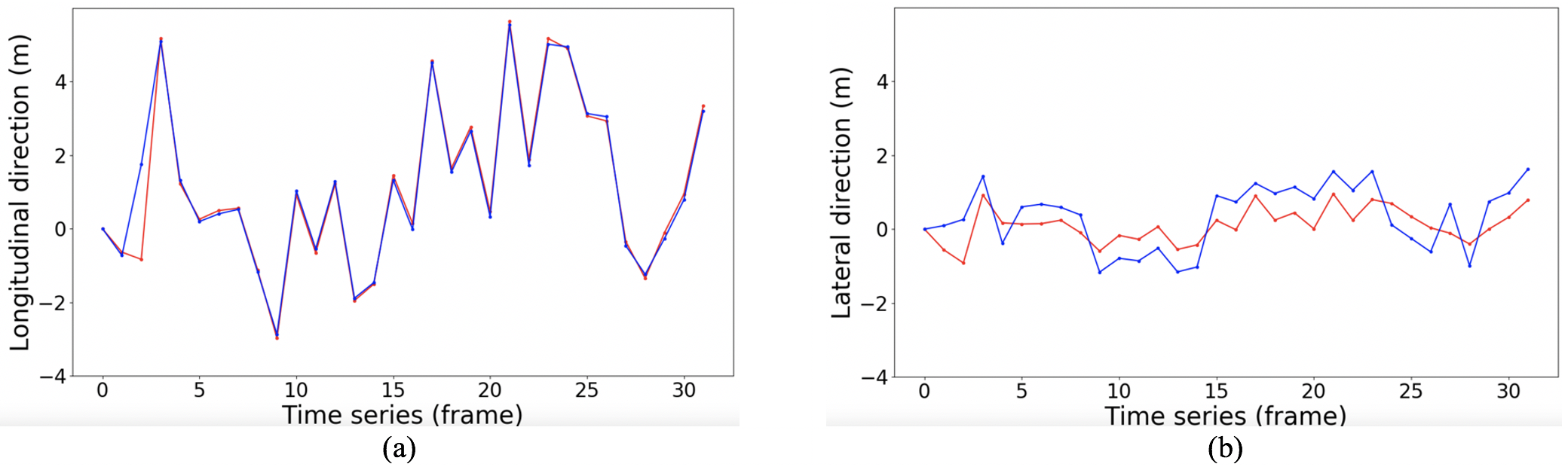}
  \caption{The deviation between the ground truth and the estimated geolocations with Approach 1 (the red polyline) and Approach 2 (the blue polyline) in the lateral and longitudinal directions in Scenario S1. }
  \label{combxy}
\end{figure}

Although Approach 2 can estimate the target vehicle's geolocation slightly better than Approach 1 on average, it cannot always be the optimum approach. Therefore, we investigated the deviations between the ground truth and estimated geolocations in the longitudinal and lateral directions in time series using both approaches. Our findings related to Scenario 1 as an example are presented in Figure  \ref{combxy}.

As Figure \ref{combxy} shows, the deviations between the ground truth and the estimated geolocations in the longitudinal direction (a) with Approach 1 and Approach 2 are almost overlapped. However, this parameter in the lateral direction (b) is more different. This result may be explained by the fact that the ego vehicle's geolocation was utilized in estimating the angle between the ego vehicle and the target vehicle. However, the used ego vehicle's geolocation is not noise-free. In addition, the ground truth data to apply the estimation deviation were collected by such GPS receivers, as well. Another possible explanation for this is that the employed methodology to estimate the distance between ego vehicle and target vehicle with both approaches were different. The first approach relied on the accuracy of the bounding box added by YOLO-v3 around the target vehicle, and the second approach used the central point on the bottom edge of the bounding box. In general, by considering the proposed methodology in each approach, we can conclude in case the lane marks and the vanishing point, which are needed by the second approach, are available, we can use Approach 2 as it is identified as the more accurate approach; otherwise, Approach 1 can be used. In both cases, enhancing the accuracy of the GPS receiver and the vehicle detection algorithms are vital. 

\section{Discussion }
To date, most of the existing studies on collecting traffic data focused on two main approaches: (1) stationary sensors and (2) V2V and V2I communications, which require most of the vehicles to have sensing and communication capabilities. However, the much-debated question is how to estimate the geolocation of the target vehicle in mixed traffic. To address this gap, this study proposed two approaches to dynamically estimating the target vehicle's geolocation based on an ego vehicle's vision capability.

\subsection{Comparison with related work}

As presented in Section 2, most of the existing studies regarding estimating the position of the target vehicle focused on the relative target vehicle's location estimation (e.g.,  \cite{ifthekhar2015stereo}\cite{giesbrecht2009vision}) and inter-vehicle distance estimation (e.g., \cite{lee2018intervehicle}\cite{huang2017vehicle}). Despite the importance of the target-vehicle localization, there remains a paucity of empirical studies on estimating the target vehicle's geolocation in a GPS coordinate system in order to enhance the ITMS awareness about the traffic scene. In addition, to be able to generalize the proposed approaches in reality, using real traffic data to run experiments is vital; however, most of the studies applied the experiments by using a simulator (e.g., \cite{ifthekhar2015stereo}). Moreover, studying the real traffic data from the urban area, which can reflect the possible estimation uncertainties and sources of noise (e.g., tall buildings that affect the GPS receiver accuracy) are important.

Therefore, the main objective of this study is to go beyond the target vehicle localization and estimate the relative angles besides estimating distance and ego vehicle's geolocation to be able to estimate the geolocation of the target vehicle dynamically. To estimate the ego vehicle's geolocation based on a low-cost GPS receiver, we used the proposed approach in \cite{namazi2021gps}. To estimate distance and relative angle, we proposed two new approaches based on object detection and image processing and geometric computation. To assess our proposed approaches, we developed a new system and ran experiments on real traffic data collected from the urban area.

The experiments on estimating the target vehicle's geolocation showed that the estimation deviation with Approach 1 was on average between 1.38 m to 3.54 m. The results with Approach 2 were between 1.4 m and 3.51 m. In our study, the ego vehicle's speed was between 16.98 km/h and 38.04 km/h, the target vehicle's speed was between 13.64 km/h and 34.35 km/h, and the distance between the ego vehicle and the target vehicle varied between 8.81 m and 44.14 m. Comparison of our findings with those of similar studies focused on estimating the distance to the target vehicle (e.g., \cite{giesbrecht2009vision}) confirms that our geolocation estimation deviation is reasonable. For example, the approach proposed by Giesbrecht et al. \cite{giesbrecht2009vision} yielded an estimated distance with a mean and maximum error of 0.72 m and 2.42 m, respectively, with a follower speed between 7.6 km/h and 10.2 km/h and a follower separation between 10.46 m and 23.71 m. However, our experiments were applied to real traffic data collected from the urban area by considering the higher speed and distance.

\subsection{Limitations of our proposed approaches}

Although the experiments confirmed that both our proposed approaches were able to estimate the geolocation of the target vehicle accurately on the right lane of the road, each approach has some pros and cons. For instance, Approach 1 is tightly connected to the vehicle detection accuracy because we employed the size of the bounding box added by YOLO around the target vehicle to estimate its distance. However, during the experiments, we observed that these bounding boxes were shaking during the trajectory, and the size of them varies between frames, which can affect the accuracy of the distance estimation.

Approach 2 has some limitations as well. For instance, like Approach 1, Approach 2 is tightly dependent on the accuracy of the central point on the bottom edge of the bounding box around the target vehicle, which is used to transform 2D space into a 3D space. Therefore, it has a direct effect on estimating the distance of the target vehicle. In addition, the camera's pitch angle  was considered in Approach 2 to enhance the localization accuracy. To estimate the camera's pitch angle, we used a vanishing point based on the parallel lane on the road. Therefore, enhancing the lane detection accuracy would be beneficial for enhancing the accuracy. Furthermore, this pitch angle was caused by our manual installation of the camera with a suction cup on the vehicle; therefore, installing the camera more precisely is highly recommended.

In addition, in the both Approaches, to estimate the angle, we used the ego vehicle's movement direction and the central point on the bottom edge of the bounding box around the target vehicle. Therefore, the estimation error of each of these parameters has a negative effect on the angle estimation accuracy. As the ego vehicle's movement direction was estimated based on the ego vehicle's GPS data, if the GPS data is noisy, the estimation deviation will increase. The same is true for the central point on the bottom edge of the bounding box around the target vehicle, which is estimated by YOLO. The experiments showed that the central point was not stable and was shaking between frames. Therefore, to mitigate the estimation deviation, enhancing the accuracy of the YOLO and GPS data is needed. 

Another source of uncertainty is that the noisy low-cost GPS receiver used to collect the ground truth data in a metropolitan area surrounded by many tall buildings. Although we applied data pre-processing to mitigate the GPS receiver noise, it is still not noise-free. Moreover, since the study was limited to vehicle movements along straight streets with limited scenarios, more studies are needed to be able to generalize our proposed approaches by considering various vehicle movements scenarios.

\section{Conclusion}
The main goal of the current study was to dynamically estimate the geolocation of mobile target vehicles via a low-cost front-facing monocular camera on a mobile ego vehicle. To estimate the target vehicle's geolocation, the distance between the ego vehicle and the target vehicle and the relative angle are needed. In this regard, we proposed two approaches: (1) object detection and image processing and (2) geometric computation. 

The results of the evaluation using real traffic data confirmed that our algorithms were able to estimate the geolocation of the target vehicles accurately. Taken together, these findings confirmed the feasibility of a vehicle-mounted monocular camera for estimating the location of target vehicles in mixed traffic. The present study lays the groundwork for future research on using an ego vehicle as a mobile sensor to collect traffic data to reduce the traffic cost and improve ITMS efficiency. Further studies that take these data types into account will be needed to increase the accuracy and enhance the generalizability by considering various scenarios.

\bibliography{elsarticle-template}

\begin{thebibliography}{10}
\expandafter\ifx\csname url\endcsname\relax
  \def\url#1{\texttt{#1}}\fi
\expandafter\ifx\csname urlprefix\endcsname\relax\def\urlprefix{URL }\fi
\expandafter\ifx\csname href\endcsname\relax
  \def\href#1#2{#2} \def\path#1{#1}\fi

\bibitem{namazi2019slr}
E.~Namazi, J.~Li, C.~Lu, Intelligent intersection management systems
  considering autonomous vehicles: A systematic literature review, IEEE Access
  7 (2019) 91946--91965.

\bibitem{oliveira2010vehicle}
H.~A. Oliveira, F.~R. Barbosa, O.~M. Almeida, A.~P. Braga, A vehicle
  classification based on inductive loop detectors using artificial neural
  networks, in: 2010 9th IEEE/IAS International Conference on Industry
  Applications-INDUSCON 2010, IEEE, 2010, pp. 1--6.

\bibitem{kurdi2014review}
H.~A. Kurdi, Review of closed circuit television (cctv) techniques for vehicles
  traffic management, International Journal of Computer Science \& Information
  Technology 6~(2) (2014) 199.

\bibitem{coststatistics}
B.~A. Kiisa~Wiegand, Challenges of the day-today operation of a traffic
  monitoring program (2016).

\bibitem{ahangar2021survey}
M.~N. Ahangar, Q.~Z. Ahmed, F.~A. Khan, M.~Hafeez, A survey of autonomous
  vehicles: enabling communication technologies and challenges, Sensors 21~(3)
  (2021) 706.

\bibitem{SAE}
\href{https://www.sae.org/news/press-room/2018/12/sae-international-releases-updated-visual-chart-for-its-\%E2\%80\%9Clevels-of-driving-automation\%E2\%80\%9D-standard-for-self-driving-vehicles}{Sae
  international releases updated visual chart for its “levels of driving
  automation” standard for self-driving vehicles}.
\newline\urlprefix\url{https://www.sae.org/news/press-room/2018/12/sae-international-releases-updated-visual-chart-for-its-\%E2\%80\%9Clevels-of-driving-automation\%E2\%80\%9D-standard-for-self-driving-vehicles}

\bibitem{raj2020multicriteria}
A.~Raj, J.~A. Kumar, P.~Bansal, A multicriteria decision making approach to
  study barriers to the adoption of autonomous vehicles, Transportation
  research part A: policy and practice 133 (2020) 122--137.

\bibitem{namazi2019master}
E.~Namazi, R.~N. Holthe-Berg, C.~S. Lofsberg, J.~Li, Using vehicle-mounted
  camera to collect information for managing mixed traffic, in: 2019 15th
  International Conference on Signal-Image Technology \& Internet-Based Systems
  (SITIS), IEEE, 2019, pp. 222--230.

\bibitem{namazi2020lane}
E.~Namazi, J.~Li, R.~Mester, C.~Lu, Identifying and counting vehicles in
  multiple lanes by using a low-cost vehicle-mounted sensor for intelligent
  traffic management systems, in: International Conference on Hybrid Artificial
  Intelligence Systems, Springer, 2020, pp. 598--611.

\bibitem{distance}
Movable-Type,
  \href{https://www.movable-type.co.uk/scripts/latlong.html}{Calculate
  distance, bearing and more between latitude/longitude points}.
\newline\urlprefix\url{https://www.movable-type.co.uk/scripts/latlong.html}

\bibitem{javanmardi2019autonomous}
E.~Javanmardi, Y.~Gu, M.~Javanmardi, S.~Kamijo, Autonomous vehicle
  self-localization based on abstract map and multi-channel lidar in urban
  area, IATSS research 43~(1) (2019) 1--13.

\bibitem{chehri2019survey}
A.~Chehri, N.~Quadar, R.~Saadane, Survey on localization methods for autonomous
  vehicles in smart cities, in: Proceedings of the 4th International Conference
  on Smart City Applications, 2019, pp. 1--6.

\bibitem{gpsac}
N.~Acosta, J.~Toloza, Techniques to improve the gps precision, International
  Journal of Advanced Computer Science and Applications 3~(8).

\bibitem{huang2021survey}
Z.~Huang, S.~Qiao, N.~Han, C.-a. Yuan, X.~Song, Y.~Xiao, Survey on vehicle map
  matching techniques, CAAI Transactions on Intelligence Technology 6~(1)
  (2021) 55--71.

\bibitem{ifthekhar2015stereo}
M.~S. Ifthekhar, N.~Saha, Y.~M. Jang, Stereo-vision-based cooperative-vehicle
  positioning using occ and neural networks, Optics Communications 352 (2015)
  166--180.

\bibitem{hayakawa2019ego}
J.~Hayakawa, B.~Dariush, Ego-motion and surrounding vehicle state estimation
  using a monocular camera, in: 2019 IEEE Intelligent Vehicles Symposium (IV),
  IEEE, 2019, pp. 2550--2556.

\bibitem{lee2018intervehicle}
J.~Lee, Intervehicle distance estimation through camera images, Journal of
  Electronic Imaging 27~(6) (2018) 063001.

\bibitem{huang2017vehicle}
D.-Y. Huang, C.-H. Chen, T.-Y. Chen, W.-C. Hu, K.-W. Feng, Vehicle detection
  and inter-vehicle distance estimation using single-lens video camera on
  urban/suburb roads, Journal of Visual Communication and Image Representation
  46 (2017) 250--259.

\bibitem{giesbrecht2009vision}
J.~L. Giesbrecht, H.~K. Goi, T.~D. Barfoot, B.~A. Francis, A vision-based
  robotic follower vehicle, in: Unmanned Systems Technology XI, Vol. 7332,
  International Society for Optics and Photonics, 2009, p. 73321O.

\bibitem{namazi2021gps}
E.~Namazi, R.~Mester, C.~Lu, M.~M. Log, J.~Li, Improving vehicle localization
  with two low-cost gps receivers, in: The sixth smart city applications
  international conference, 2021.

\bibitem{redmon2016you}
J.~Redmon, S.~Divvala, R.~Girshick, A.~Farhadi, You only look once: Unified,
  real-time object detection, in: Proceedings of the IEEE conference on
  computer vision and pattern recognition, 2016, pp. 779--788.

\bibitem{redmon2018yolov3}
J.~Redmon, A.~Farhadi, Yolov3: An incremental improvement, arXiv preprint
  arXiv:1804.02767.

\bibitem{yoloweb}
\href{https://pjreddie.com/darknet/yolo/}{Yolo: Real-time object detection}.
\newline\urlprefix\url{https://pjreddie.com/darknet/yolo/}

\bibitem{wang2021daedalus}
D.~Wang, C.~Li, S.~Wen, Q.-L. Han, S.~Nepal, X.~Zhang, Y.~Xiang, Daedalus:
  Breaking nonmaximum suppression in object detection via adversarial examples,
  IEEE Transactions on Cybernetics.

\bibitem{kitti}
J.~Redmon,
  \href{https://github.com/packyan/PyTorch-YOLOv3-kitti}{Pytorch-yolov3-kitti}.
\newline\urlprefix\url{https://github.com/packyan/PyTorch-YOLOv3-kitti}

\bibitem{FOV}
Gopro,
  \href{https://gopro.com/help/articles/question_answer/hero7-field-of-view-fov-information?sf96748270=1}{Hero7
  field of view (fov) information}.
\newline\urlprefix\url{https://gopro.com/help/articles/question_answer/hero7-field-of-view-fov-information?sf96748270=1}

\bibitem{ding2001canny}
L.~Ding, A.~Goshtasby, On the canny edge detector, Pattern Recognition 34~(3)
  (2001) 721--725.

\bibitem{galamhos1999progressive}
C.~Galamhos, J.~Matas, J.~Kittler, Progressive probabilistic hough transform
  for line detection, in: Proceedings. 1999 IEEE computer society conference on
  computer vision and pattern recognition (Cat. No PR00149), Vol.~1, IEEE,
  1999, pp. 554--560.

\bibitem{matas2000robust}
J.~Matas, C.~Galambos, J.~Kittler, Robust detection of lines using the
  progressive probabilistic hough transform, Computer vision and image
  understanding 78~(1) (2000) 119--137.

\bibitem{thalestheorem}
\href{https://sites.math.washington.edu/~king/coursedir/m444a02/class/10-21-thales.html}{Thales
  theorem and figure}.
\newline\urlprefix\url{https://sites.math.washington.edu/~king/coursedir/m444a02/class/10-21-thales.html}

\bibitem{Euclideandis}
\href{https://en.wikipedia.org/wiki/Euclidean_distance}{Euclidean distance}.
\newline\urlprefix\url{https://en.wikipedia.org/wiki/Euclidean_distance}

\bibitem{Trigonometry}
\href{https://en.wikipedia.org/wiki/Trigonometry}{Trigonometry}.
\newline\urlprefix\url{https://en.wikipedia.org/wiki/Trigonometry}

\end{thebibliography}

\newpage

\end{document}